\documentstyle[aclap]{article}

\title{An Information Structural Approach\\to Spoken Language Generation}

\author{Scott Prevost \\
The Media Laboratory \\
Massachusetts Institute of Technology \\
20 Ames Street \\
Cambridge, Massachusetts 02139-4307 USA \\
{\tt prevost@media.mit.edu}}

\input{psfig}

\newtheorem{example}{}

\newcommand{\startx} 
 {
  \begin{example} \rm \ \begin{minipage}[t]{.48\textwidth}}

\newcommand{\startxl}[1] 
  {
  \begin{example} \label{#1} \ \rm \begin{minipage}[t]{.4\textwidth}}

\newcommand{\startpx} 
 {
  \begin{example} \rm \ \begin{minipage}[t]{.48\textwidth}}

\newcommand{\startpxl}[1] 
 {
  \begin{example} \label{#1} \ \rm \begin{minipage}[t]{.48\textwidth}}

\newcommand{\stopx} 
 {\end{minipage} \end{example} 
        \vspace{\jot}}

\newcommand{\stopxcont} 
 {\end{minipage} \end{example} \vspace{-\belowdisplayskip}
        \vspace{\jot}}

\newcommand{\startfxl}[1] 
 {\begin{figure*}[p]
  \vspace{-.5in}
  \begin{example}
  \label{#1} \rm \
  \begin{minipage}[t]{\textwidth}}

\newcommand{\startfxll}[1] 
 {\begin{figure*}
  \begin{example}
  \label{#1} \rm \
  \begin{minipage}[t]{.9\textwidth}}

\newcommand{\startfxlt}[1] 
 {\begin{figure*}[t]
  \begin{example}
  \label{#1} \rm \
  \begin{minipage}[t]{\textwidth}}

\newcommand{\stopfx} 
 {\stopx \end{figure*}}

\newcommand{\startwxl}[1]       
  {
  \begin{example}
  \label{#1} \ \rm
  \begin{minipage}[t]{\textwidth}}

\newenvironment{lexlist}
{\begin{list}{}
{\setlength{\itemsep}{0ex}
\setlength{\parsep}{\parskip}
\setlength{\leftmargin}{1.5em}
\setlength{\labelsep}{.5em}
\setlength{\labelwidth}{2em}}}{\end{list}}

\newcommand{\tb}[1]{\begin{tabbing} #1 \end{tabbing}}

\newcommand{\annotate}[1]{
	\begin{tabular}[t]{c} #1 \end{tabular}}

\newcommand{\rannotate}[1]{
	\begin{tabular}[t]{r} #1 \end{tabular}}

\newcommand{\boxannotate}[1]{
	\fbox{\begin{tabular}[t]{c} #1 \end{tabular}}}

\begin{document}
\maketitle
\bibliographystyle{fullname}

\vspace{-0.5in}
\begin{abstract}
This paper presents an architecture for the generation of 
spoken monologues with contextually appropriate intonation. 
A two-tiered {\em information structure} 
representation is used in the high-level content planning and
sentence planning stages of generation to produce efficient, coherent
speech that makes certain discourse relationships, such as explicit
contrasts, appropriately salient.  The system
is able to produce appropriate intonational patterns that cannot be
generated by other systems which rely solely on word class and given/new
distinctions.
\end{abstract}

\section{Introduction}

While research on generating coherent {\em written} text has
flourished within the computational linguistics and artificial
intelligence communities, research on the generation of {\em spoken}
language, and particularly intonation, has received somewhat less
attention.  In this paper, we argue that commonly employed models of
text organization, such as schemata and rhetorical structure theory
(RST), do not adequately address many of the issues involved in
generating {\em spoken} language.
Such approaches fail to consider contextually bound
focal distinctions that are manifest through a variety of different
linguistic and paralinguistic devices, depending on the language.

In order to account for such distinctions of focus, we employ a
two-tiered {\em information structure} representation as a framework
for maintaining local coherence in the generation of natural language.
The higher tier, which delineates the {\em theme}, that which links
the utterance to prior utterances, and the {\em rheme}, that which
forms the core contribution of the utterance to the discourse, is
instrumental in determining the high-level organization of information
within a discourse segment.  Dividing semantic
representations into their thematic and rhematic parts 
allows propositions to be presented in a way that
maximizes the shared material between utterances.

The lower tier in the information structure representation specifies
the semantic material that is in ``focus'' within themes and rhemes.
Material may be in focus for a variety of reasons, such as to
emphasize its ``new'' status in the discourse, or to contrast it with
other salient material.  Such focal distinctions may affect the
linguistic presentation of information.  For example, the it-cleft
in~\ref{ex:itcleft} may mark {\em John} as standing in contrast to
some other recently mentioned person.  Similarly, in~\ref{ex:accent},
the pitch accent on {\em red} may mark the referenced car as standing
in contrast to some other car inferable from the discourse
context.\footnote{ In this example, and throughout the remainder of the
paper, the intonation contour is informally noted by placing prosodic
phrases in parentheses and marking pitch accented words with capital
letters.  The tunes are more formally annotated with a variant of
\cite{pier80} notation described in 
\cite{prev95}.  Three different pause lengths are associated with
boundaries in the modified notation.  '(\%)' marks intra-utterance
boundaries with very little pausing, '\%' marks intra-utterance
boundaries associated with clauses demarcated by commas, and '\$'
marks utterance-final boundaries.  For the purposes of generation and
synthesis, these distinctions are crucial.}

\startxl{ex:itcleft}
It was John who spoke first.
\stopxcont
\startxl{ex:accent}
\begin{lexlist}
\item [Q:] Which car did Mary drive?
\item [A:] \tb{(\={\sc mary} drove\=)$_{th}$\hspace{1cm} 
(the R\=ED car\=.)$_{rh}$\\
\>L+H* \> LH(\%) \> H* \> LL\$}
\end{lexlist}
\stopx

\startfxll{ex:minpair1}
\begin{lexlist}
\small
\raggedright
\item [Q:] I know the AMERICAN amplifier produces MUDDY treble,\\
(But \annotate{WHAT)\\L+H* L(H\%)} 
(does the \annotate{BRITISH\\H*} amplifier \rannotate{produce?)\\LL\$}
\item [A:]
\boxannotate{
   (The \boxannotate{BRITISH\\L+H*\\{\em theme-focus}} amplifier 
\rannotate{produces)\\L(H\%)}\\{\em Theme}}
\boxannotate{
   \boxannotate{(CLEAN\\H*\\{\em rheme-focus}} \rannotate{treble.)\\LL\$}\\
   {\em Rheme}}
\end{lexlist}
\stopfx

\startfxll{ex:minpair2}
\begin{lexlist}
\small
\raggedright
\item [Q:] I know the AMERICAN amplifier produces MUDDY treble,\\
(But \annotate{WHAT)\\L+H* L(H\%)}
(produces \annotate{CLEAN\\H*} \rannotate{treble?)\\LL\$}
\item [A:]
\boxannotate{
   (The \boxannotate{BRITISH\\H*\\{\em rheme-focus}} 
   \rannotate{amplifier)\\L(L\%)} \\ {\em Rheme}}
\boxannotate{
   (produces \boxannotate{CLEAN\\L+H*\\{\em theme-focus}} 
   \rannotate{treble.)\\LH\$}\\ {\em Theme}}
\end{lexlist}
\stopfx

By appealing to the notion that the simple rise-fall tune (H* LL\%)
very often accompanies the rhematic material in an utterance and the
rise-fall-rise tune often accompanies the thematic material 
\cite{stee91,prev94}, we
present a spoken language generation architecture for producing short
spoken monologues with contextually appropriate intonation.

\section{Information Structure\label{sec:info_struc}}

{\em Information Structure} refers to the organization of information
within an utterance.  In particular, it defines how the information
conveyed by a sentence is related to the knowledge of the
interlocutors and the structure of their discourse.  Sentences
conveying the same propositional content in different contexts need
not share the same information structure.  
That is, information structure refers to how the semantic content of an
utterance is packaged, and amounts to instructions for updating the
models of the discourse participants.  The realization of
information structure in a sentence, however, differs from language to
language.  In English, for example, intonation carries much of the
burden of information structure, while languages with freer word
order, such as Catalan \cite{engd94} and Turkish
\cite{hoff95} convey information structure syntactically.

\subsection{Information Structure and Intonation}

The relationship between intonational structure and information
structure is illustrated by~\ref{ex:minpair1} and
\ref{ex:minpair2}.  In each of these examples, the answer contains the 
same string words but different intonational patterns and information
structural representations.  The theme of each utterance is considered
to be represented by the material repeated from the question.  That
is, the theme of the answer is what links it to the question and
defines what the utterance is about.  The rheme of each utterance is
considered to be represented by the material that is new or forms the
core contribution of the utterance to the discourse.  By mapping the
rise-fall tune (H* LL\%) onto rhemes and the rise-fall-rise tune (L+H*
LH\%) onto themes \cite{stee91,prev94}, we can easily identify the
string of words over which these two prominent tunes occur directly from 
the information structure.  While this mapping is certainly
overly simplistic, the results presented in
Section~\ref{sec:results} demonstrate its appropriateness for the
class of simple declarative sentences under investigation.

Knowing the strings of words to which these two tunes are to be
assigned, however, does not provide enough information to determine
the location of the pitch accents (H* and L+H*) within the tunes.
Moreover, the simple mapping described above does not account for the
frequently occurring cases in which thematic material bears no pitch 
accents and is consequently unmarked intonationally.
Previous approaches to the problem of determining where to place
accents have utilized heuristics based on ``givenness.''  That
is, content-bearing words (e.g. nouns and verbs) which had not been
previously mentioned (or whose roots had not been previously
mentioned) were assigned accents, while function words were de-accented
\cite{davi88,hirs90}.  While these heuristics account for a
broad range of intonational possibilities, they fail to account for
accentual patterns that serve to contrast entities or propositions
that were previously ``given'' in the discourse.  Consider, for
example the intonational pattern in~\ref{ex:contrast1}, in which the
pitch accent on {\em amplifier} in the response cannot be attributed
to its being ``new'' to the discourse.

\startxl{ex:contrast1}
\begin{lexlist}
\item [Q:] \tb{Do critics prefer the BR\=ITISH amplifi\=er\\
\>L* \> H \\
or the AM\=ERICAN amplifie\=r?\\
\> H* \> LL\$}
\item [A:] \tb{They prefer
the AM\=ERICAN amplif\=ier.\\
\> H* \> LL\$}
\end{lexlist}
\stopx

For the determination of pitch accent placement, we rely on a
secondary tier of information structure which identifies focused
properties within themes and rhemes.  The theme-foci and the
rheme-foci mark the information that differentiates properties or
entities in the current utterance from properties or entities
established in prior utterances.  Consequently, the semantic material
bearing ``new'' information is considered to be in focus.
Furthermore, the focus may include semantic material that serves to
contrast an entity or proposition from alternative entities or
propositions already established in the discourse.  While the types of
pitch accents (H* or L+H*) are determined by the theme/rheme
delineation and the aforementioned mapping onto tunes, the locations
of pitch accents are determined by the assignment of foci within the
theme and rheme, as illustrated in~\ref{ex:minpair1} and
\ref{ex:minpair2}.  Note that it is in precisely those cases where
thematic material, which is ``given'' by default, does not
contrast with any other previously established properties or entities
that this material is intonationally unmarked, as in
\ref{ex:unmarked}.

\startxl{ex:unmarked}
\begin{lexlist}
\item [Q:] \tb{Which amplifier does Scott PRE\=FER?\\
\> H* LL\$}
\item [A:] \tb{(He prefers)$_{\em th}$  (the BR\=ITISH
amplifie\=r.)$_{\em rh}$\\
\> H* \> LL\$}
\end{lexlist}
\stopx

\subsection{Contrastive Focus Algorithm}

The determination of contrastive focus, and consequently the determination
of pitch accent locations, is based on the premise
that each object in the knowledge base is associated with a set of
alternatives from which it must be distinguished if reference is to
succeed.  The set of alternatives is determined by the hierarchical
structure of the knowledge base.  For the present implementation, only
properties with the same parent or grandparent class are considered to
be alternatives to one another.


Given an entity $x$ and a referring expression for $x$, the
contrastive focus feature for its semantic representation is computed
on the basis of the contrastive focus algorithm described in~\ref{ex:notation},
\ref{ex:initial} and \ref{ex:pseudo}. 
The data structures and notational conventions are given below.

\startxl{ex:notation}
\begin{description}
\item[{\em DElist}:] a collection of discourse entities that have
been evoked in prior discourse, ordered by recency.  The list may be
limited to some size $k$ so that only the $k$ most recent discourse
entities pushed onto the list are retrievable.
\item[{\em ASet}$(x)$:] the set of alternatives for object $x$, i.e. those
objects that belong to the same class as $x$, as defined in the
knowledge base.
\item[{\em RSet}$(x,S)$:]  the set of alternatives for object $x$ as 
restricted by the referring expressions in {\em DElist} and the set of 
properties $S$.
\item[{\em CSet}$(x,S)$:] the subset of properties of $S$ to be accented 
for contrastive purposes.
\item[{\em Props}$(x)$:] a list of properties for object $x$, ordered by the
grammar so that
nominal properties take precedence over adjectival
properties.
\end{description}
\stopx

The algorithm, which assigns contrastive focus in
both thematic and rhematic constituents, begins by isolating the
discourse entities in the given constituent.  For each
such entity 
$x$, the structures defined above are initialized as follows:
\startxl{ex:initial}
\begin{description}
\small
\item[{\em Props}$(x)$ :=]  $[P$ $|$ $P(x)$ is true in KB $]$
\item[{\em ASet}$(x)$ :=] $\{y$ $|$ {\em alt}$(x,y)\}$, $x$'s alternatives
\item[{\em RSet}$(x,\{\})$ :=] $\{x\} \cup \{y$ $|$ $y \in $\ {\em ASet}$(x)
\ \&\ y \in $\ {\em DElist}$\}$, evoked alternatives
\item[{\em CSet}$(x,\{\})$ :=] $\{\}$\\
\end{description}
\stopx
The algorithm appears in pseudo-code in~\ref{ex:pseudo}.\footnote{An 
in-depth discussion of the algorithm and numerous examples are presented 
in \cite{prev95}.}
\startxl{ex:pseudo}
\begin{tabbing}
$S$ \=:= \=$\{\}$\\
{\bf for each} $P$ {\bf in} {\em Props}$(x)$\\
\>	{\em RSet}$(x, S \cup \{P\})$ :=\\
\>\> $\{y$ $|$ $y \in $\ {\em RSet}$(x,S)$ $\&$ $P(y)\}$\\
\>	{\bf if} {\em RSet}$(x,S \cup \{P\}) = $\ {\em RSet}$(x,S)$ 
{\bf then}\\
\> \>		\% no restrictions were made\\
\> \>		\% based on property $P$.\\
\> \>		{\em CSet}$(x,S \cup \{P\}) := $\ {\em CSet}$(x,S)$\\
\>	{\bf else}\\
\> \>		\% property $P$ eliminated some\\
\> \>		\% members of the {\em RSet}.\\
\> \>		{\em CSet}$(x,S \cup \{P\}) := $\ {\em CSet}$(x,S) \cup 
\{P\}$\\
\>	{\bf endif}\\
\>	$S := S \cup \{P\}$\\
{\bf endfor}\\
\end{tabbing}
\stopx

In other words, given an object $x$, a list of its properties and a
set of alternatives, the set of alternatives is restricted by
including in the initial {\em RSet} only $x$ and those objects that
are explicitly referenced in the prior discourse.  Initially, the set
of properties to be contrasted ({\em CSet}) is empty.  Then, for each
property of $x$ in turn, the {\em RSet} is restricted to include only
those objects satisfying the given property in the knowledge base.  If
imposing this restriction on the {\em RSet} for a given property
decreases the cardinality of the {\em RSet}, then the property serves
to distinguish $x$ from other salient alternatives evoked in the prior
discourse, and is therefore added to the contrast set.  Conversely, if
imposing the restriction on the {\em RSet} for a given property does
not change the {\em RSet}, the property is {\em not} necessary for
distinguishing $x$ from its alternatives, and is not added to the {\em
CSet}.

Based on this contrastive focus algorithm and the mapping between
information structure and intonation described above, we can view
information structure as the representational bridge between discourse
and intonational variability.  The following sections elucidate how
such a formalism can be integrated into the computational task of
generating spoken language.


\section{Generation Architecture}\label{gen_frame}

The task of natural language generation (NLG) has often been divided
into three stages: content planning, in which high-level goals are
satisfied and discourse structure is determined, sentence planning, in
which high-level abstract semantic representations are mapped onto
representations that more fully constrain the possible sentential
realizations \cite{ramb92,reit92b,mete91}, and surface generation,
in which the high-level propositions are converted into sentences.

The selection and organization of propositions and their divisions into
theme and rheme are determined by the content planner, which maintains 
discourse coherence by stipulating that semantic information must be 
shared between
consecutive utterances whenever possible.  That is, the content planner
ensures that the theme of an utterance links it to material in prior
utterances.

The process of determining foci within themes and rhemes can be
divided into two tasks: determining which discourse {\em entities} or
{\em propositions} are in focus, and determining how their linguistic
realizations should be marked to convey that focus.  The first of
these tasks can be handled in the content phase of the NLG model
described above.  The second of these tasks, however, relies on
information, such as the construction of referring expressions, that
is often considered the domain of the sentence planning stage.  For
example, although two discourse entities $e_1$ and $e_2$ can be
determined to stand in contrast to one another by appealing only to
the discourse model and the salient pool of knowledge, the method of
contrastively distinguishing between them by the
placement of pitch accents cannot be resolved until the choice of
referring expressions has been made.  
Since referring expressions are generally taken to be in the domain 
of the sentence planner 
\cite{dale91}, the present approach resolves issues of
contrastive focus assignment at the sentence processing stage as well.

During the content generation phase, the content of the utterance is
planned based on the previous discourse.  While schema-based systems
\cite{mcke85} have been widely used, rhetorical structure theory (RST)
approaches \cite{mann86,hovy93}, which organize texts by identifying
rhetorical relations between clause-level propositions from a
knowledge base, have recently flourished.  Sibun
\cite{sibu91} offers yet another alternative in which
propositions are linked to one another not by rhetorical relations or
pre-planned schemata, but rather by physical and spatial properties
represented in the knowledge-base.

The present framework for organizing the content of a monologue is a
hybrid of the schema and RST approaches.  The implementation,
which is presented in the following section, produces descriptions of
objects from a knowledge base with context-appropriate intonation that
makes proper distinctions of contrast between alternative, salient
discourse entities.  Certain constraints, such as the requirement that
objects be identified or defined at the beginning of a description,
are reminiscent of McKeown's schemata.  Rather than imposing strict
rules on the order in which information is presented, the order is
determined by domain specific knowledge, the communicative intentions
of the speaker, and beliefs about the hearer's knowledge.  Finally,
the system includes a set of rhetorical constraints that may rearrange
the order of presentation for information in order to make certain
rhetorical relationships salient.  While this approach has proven
effective in the present implementation, further research is required
to determine its usefulness for a broader range of discourse types.

\section{The Prolog Implementation}\label{monologue}

The monologue generation program 
produces text and contextually-appropriate intonation contours
to describe an object from the knowledge base.  The
system exhibits the ability to intonationally contrast alternative
entities and properties that have been explicitly evoked in the
discourse even when they occur with several intervening sentences.

\subsection{Content Generation}

\begin{figure}[bt]
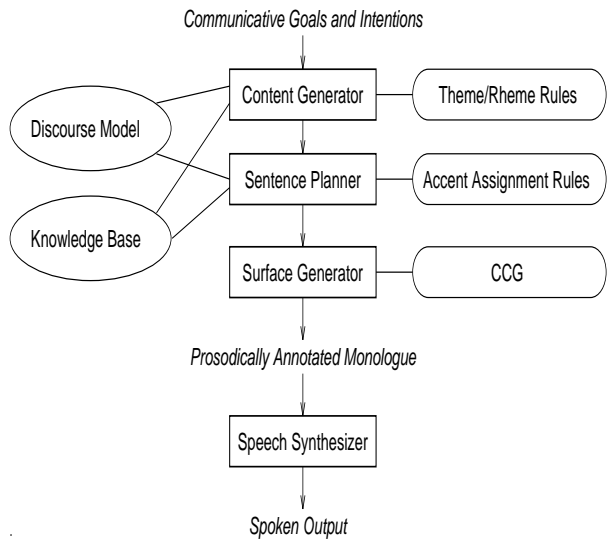

\centerline{\psfig{figure=arch.eps,{width={.48\textwidth}},{height={7cm}}}}
\caption{An Architecture for Monologue Generation}
\label{ex:arch2}
\end{figure}

The architecture for the monologue generation program is shown in
Figure~\ref{ex:arch2}, in which arrows represent the computational flow
and lines represent dependencies among modules.  The remainder of this
section contains a description of the computational path through the
system with respect to a single example.  The input to the program is
a goal to describe an object from the knowledge base, which in this
case contains a variety of facts about hypothetical stereo
components. In addition, the input provides a
communicative intention for the goal which may affect its ultimate
realization, as shown in \ref{ex:mono1}.  For example, given the goal
{\tt describe(x)}, the intention {\tt persuade-to-buy(hearer,x)} may
result in a radically different monologue than the intention {\tt
persuade-to-sell(hearer,x)}.

\startxl{ex:mono1}
\begin{description}
\item[{\em Goal:}]  {\tt describe e1}
\item[{\em Input:}]  {\tt generate(intention(bel(h1, good-to-buy(e1)))}
\end{description}
\stopx

Information from the knowledge base is selected to be included in the
output by a set of
relations that determines the degree to which knowledge base facts and
rules support the communicative intention of the speaker.
For example, suppose the system ``believes'' that conveying the proposition
in~\ref{ex:mono2} moderately supports the intention of making hearer
\verb#h1# want to buy \verb#e1#, and further that the rule
in~\ref{ex:mono3} is known by \verb#h1#.\

\startxl{ex:mono2}
\verb#bel(h1, holds(rating(X, powerful)))#
\stopxcont
\startxl{ex:mono3}
\begin{verbatim}
holds(rating(X, powerful)) :-
  holds(produce(X, Y)),
  holds(isa(Y, watts-per-channel)),
  holds(amount(Y, Z)),
  number(Z),
  Z >= 100.
\end{verbatim}
\stopx

The program then consults
the facts in the knowledge base, verifies that the property does
indeed hold and consequently includes the corresponding facts in the
set of properties to be conveyed to the hearer, as shown
in~\ref{ex:mono4}.

\startxl{ex:mono4}
\begin{verbatim}
holds(produce(e1,  e7)).
holds(isa(e7, watts-per-channel)).
holds(amount(e7, 100)).
\end{verbatim}
\stopx

The content generator starts with a simple description schema that
specifies that an object is to be explicitly identified or defined
before other propositions concerning it are put forth.  Other relevant
propositions concerning the object in question are then linearly
organized according to beliefs about how well they contribute to the
overall intention.  Finally, a small set of
rhetorical predicates rearranges the linear ordering of propositions
so that sets of sentences that stand in some interesting rhetorical
relationship to one another will be realized together in the output.
These rhetorical predicates employ information structure to assist in
maintaining the coherence of the output.  For example, the {\em
conjunction} predicate specifies that propositions sharing the same
theme or rheme be realized together in order to avoid excessive topic
shifting.  The {\em contrast} predicate specifies that pairs of themes
or rhemes that explicitly contrast with one another be realized
together.  The result is a set of properties roughly ordered by
the degree to which they support the given intention, as shown 
in~\ref{ex:props}.

\startxl{ex:props}
\begin{verbatim}
holds(defn(isa(e1,amplifier)))
holds(design(e1,solid-state),pres)
holds(cost(e1,e9),pres)
holds(produce(e1,e7),pres)
holds(contrast(praise(e4,e1),
               revile(e5,e1)),past)
\end{verbatim}
\stopx

The top-level propositions shown in~\ref{ex:props} were selected by
the program because the hearer ({\tt h1}) is believed to be interested
in the design of the amplifier and the reviews the amplifier has
received.  Moreover, the belief that the hearer is interested in
buying an expensive, powerful amplifier justifies including
information about its cost and power rating.  Different sets of
propositions would be generated for other (perhaps thriftier) hearers.
Additionally, note that the propositions {\tt praise(e4,e1)} and {\tt
revile(e5,e1)} are combined into the larger proposition {\tt
contrast(praise(e4,e1),revile(e5,e1))}.  This is accomplished by the
rhetorical constraints that determine the two propositions to be
contrastive because {\tt e4} and {\tt e5} belong to the same set of
alternative {\em entities} in the knowledge base and {\tt praise} and
{\tt revile} belong to the same set of alternative {\em propositions}
in the knowledge base.

The next phase of content generation recognizes the dependency
relationships between the properties to be conveyed based on shared
discourse entities.  This phase, which represents an extension of the
rhetorical constraints, arranges propositions to ensure that
consecutive utterances share semantic material (cf. \cite{mcke94}).  
This rule, which in effect imposes a
strong bias for Centering Theory's {\em continue} and {\em retain}
transitions \cite{gros86b} determines the
theme-rheme segmentation for each proposition.

\subsection{Sentence Planning}

After the coherence constraints from the previous section are applied,
the sentence planner is responsible for making decisions concerning
the form in which propositions are realized.  This is accomplished by
the following simple set of rules.  First, Definitional {\tt isa}
properties are realized by the matrix verb.  Other {\tt isa}
properties are realized by nouns or noun phrases.  Top-level
properties (such as those in~\ref{ex:props}) are realized by the
matrix verb.  Finally, embedded properties (those evoked for building
referring expressions for discourse entities) are realized by
adjectival modifiers if possible and otherwise by relative clauses.

While there are certainly a number of linguistically interesting
aspects to the sentence planner, the most important aspect for the
present purposes is the determination of theme-foci and rheme-foci.
The focus assignment algorithm employed by the sentence planner, which
has access to both the discourse model and the knowledge base, works
as follows.  First, each property
or discourse entity in the semantic and information structural
representations is marked as either previously mentioned or new to the
discourse.  This assignment is made with respect to two data structures, 
the discourse
entity list (DEList), which tracks the succession of entities through 
the discourse, and a
similar structure for evoked properties.  Certain aspects of the
semantic form are considered unaccentable because they correspond to
the interpretations of closed-class items such as function words.
Items that are assigned focus based on their ``newness''
are assigned the $\circ$ focus operator, as
shown in \ref{ex:mono7}.

\startxl{ex:mono7}
\begin{tabular}[t]{ll}
Semantics: & ${\em defn}({\em isa}({\circ}e1,{\circ}c1))$\\
Theme: & ${\circ}e1$\\
Rheme: & $\lambda x.{\em isa}(x,{\circ}c1)$\\
Supporting Props: & ${\em isa}(c1, \circ{\em amplifier})$\\
& $\circ{\em design}(c1, \circ{\em solidstate})$
\end{tabular}
\stopx

The second step in the focus assignment algorithm checks for the
presence of contrasting propositions in the ISStore, a structure that
stores a history of information structure representations.
Propositions are considered contrastive if they contain two
contrasting pairs of discourse entities, or if they contain one
contrasting pair of discourse entities as well as contrasting
functors.  

Discourse entities are determined to be contrastive if they
belong to the same set of alternatives in the knowledge base, where
such sets are inferred from the {\tt isa}-links that define class
hierarchies.  While the present
implementation only considers entities with the same parent or grandparent
class to be alternatives for the purposes of contrastive stress, a
graduated approach that entails degrees of contrastiveness may also be
possible. 

The effects of the focus assignment algorithm are easily shown by
examining the generation of an utterance that contrasts with the
utterance shown in~\ref{ex:mono7}.  That is, suppose the generation
program has finished generating the output corresponding to the
examples in \ref{ex:mono1} through \ref{ex:mono7} and is assigned the
new goal of describing entity {\tt e2}, a different amplifier.  After
applying the second step on the focus assignment algorithm,
contrasting discourse entities are marked with the $\bullet$ contrastive
focus operator,
as shown in \ref{ex:mono9}.  Since {\tt e1} and {\tt e2} are both
instances of the class {\tt amplifiers} and {\tt c1} and {\tt c2} both
describe the class {\tt amplifiers} itself, these two pairs of discourse
entities are considered to stand in contrastive relationships.

\startxl{ex:mono9}
\begin{tabular}[t]{ll}
Semantics: & ${\em defn}({\em isa}({\bullet}e2,{\bullet}c2))$\\
Theme: & ${\bullet}e2$\\
Rheme: & $\lambda x.{\em isa}(x,{\bullet}c2)$\\
Supporting Props: & ${\em class}(c2, {\em amplifier})$\\
& ${\em design}(c2, \circ{\em tube})$
\end{tabular}
\stopx

While the previous step of the algorithm determined which abstract
discourse entities and properties stand in contrast, the third step
uses the contrastive focus algorithm described in
Section~\ref{sec:info_struc} to determine which elements need to be
contrastively focused for reference to succeed.  This algorithm
determines the minimal set of properties of an entity that must be
``focused'' in order to distinguish it from other salient entities.
For example, although the representation in \ref{ex:mono9} specifies
that {\tt e2} stands in contrast to some other entity, it is the
property of {\tt e2} having a tube design rather than a solid-state
design that needs to be conveyed to the hearer.  After applying the
third step of the focus assignment to \ref{ex:mono9}, the result
appears as shown in ~\ref{ex:mono10}, with ``tube'' contrastively
focused as desired.
\startxl{ex:mono10}
\begin{tabular}[t]{ll}
Semantics: & ${\em defn}({\em isa}({\bullet}e2,{\bullet}c2))$\\
Theme: & ${\bullet}e2$\\
Rheme: & $\lambda x.{\em isa}(x,{\bullet}c2)$\\
Supporting Props: & ${\em isa}(c2, {\em amplifier})$\\
& ${\em design}(c2, \bullet{\em tube})$
\end{tabular}
\stopx

The final step in the sentence planning phase of generation is to
compute a representation that can serve as input to a surface form
generator based on Combinatory Categorial Grammar (CCG) 
\cite{stee91}, as shown in ~\ref{ex:mono12}.\footnote{
A complete description of the CCG generator
can be found in
\cite{prev93}.  CCG was chosen as the grammatical formalism because
it licenses non-traditional syntactic constituents that are congruent
with the bracketings imposed by information structure and intonational
phrasing, as illustrated in~\ref{ex:minpair1}.}

\startxl{ex:mono12}
\begin{description}
\small
\item[Theme:] $np(3,s):$\\
$(e1\verb#^#S)\verb#^#{\em def}(e1, {\bullet}x5(e1)\&S) @ u/rh$
\item[Rheme:] $s:$\\
$({\em act}\verb#^#{\em pres})\verb#^#{\em indef}(c1, ({\em
amplifier}(c1) \& $\\
${\bullet}{\em tube}(c1))\&{\em isa}(e1,c1)){\backslash}np(3,s):e1@rh$
\end{description}
\stopx

\subsection{Results\label{sec:results}}

Given the focus-marked output of the sentence planner, the surface generation
module consults a CCG grammar which encodes the information
structure/intonation mapping and
dictates the generation of both the syntactic and prosodic
constituents.  The result is a string of words and the appropriate
prosodic annotations, as shown in~\ref{ex:mono11a}.  The output of
this module is easily translated into a form suitable for a speech
synthesizer, which produces spoken output with the desired
intonation.\footnote{The system currently uses the AT\&T Bell
Laboratories TTS system, but the implementation is easily adaptable to
other synthesizers.}

\startxl{ex:mono11a}
\tb{The \=X5\hspace{1.5cm} is a T\=UBE amplifier\=.\\
\> L+H$_c^*$ L(H\%) \>H$_c^*$ \> LL\$}
\stopx

The modules described above and shown in Figure~\ref{ex:arch2} are
implemented in Quintus Prolog.  The system
produces the types of output shown in
\ref{ex:out1} and \ref{ex:out2}, which should be interpreted as
a single (two paragraph) monologue satisfying a goal to describe two
different objects.\footnote{The implementation assigns slightly higher
pitch to accents bearing the subscript $c$ (e.g. H$_c^*$), which mark
contrastive focus as determined by the algorithm describe above and
in \cite{prev95}.} Note that both paragraphs include very
similar types of information, but radically different intonational
contours, due to the discourse context.  In fact, if the intonational
patterns of the two examples are interchanged, the resulting speech
sounds highly unnatural.

\startxl{ex:out1}
\begin{lexlist}
\small
\item [a.] Describe the x4.
\item [b.] \tb{The \=X4\\
\> L+H* L(H\%) \\
\pushtabs
is a S\=OLID-state \=AMPLIF\=IER.\\
\> H* \> H* \> LL\$\\
\poptabs
\pushtabs
It C\=OSTS \=EIGHT H\=UNDRED D\=OLLARS,\\
\> H* \> H* \> H* \> H* LL\%\\
\poptabs
\pushtabs
and PROD\=UCES\\
\> H* \\
\poptabs
\pushtabs
ONE hundred watts-per-CH\=ANNEL.\\
H* \> H* LL\$\\
\poptabs
\pushtabs
It was PR\=AISED by \=STEREOF\=OOL,\\
\> H$_c^*$ \> !H$_c^*$ \> LH\%\\
\poptabs
\pushtabs
an \=AUDIO J\=OUR\=NAL,\\
\> H* \> H* \> LH\%\\
\poptabs
\pushtabs
but was REV\=ILED by  \=AUDIO\=FAD,\\
\> H$_c^*$ \> !H$_c^*$ \> LH\%\\
\poptabs
\pushtabs
AN\=OTHER audio jour\=nal.\\
\> H* \> LL\$\\
\poptabs}
\end{lexlist}
\stopxcont
\startxl{ex:out2}
\begin{lexlist}
\small
\item [a.] Describe the x5.
\item [b.] \tb{The \=X5 \hspace{1.5cm}  is a T\=UBE
amplif\=ier.\\
\> L+H$_c^*$ L(H\%) \> H$_c^*$ \> LL\$ \\
\pushtabs
\=IT\hspace{1cm} costs \=NINE hundred doll\=ars,\\
\> L+H$_c^*$ L(H\%) \> H$_c^*$ \> LH\%\\
\poptabs
\pushtabs
produces T\=WO hundred watts-per-ch\=annel.\\
\> H$_c^*$ \> LH\%\\
\poptabs
\pushtabs
and was praised\\
by Stereofool \=AND Aud\=iofad.\\
\> H$_c^*$ \> LL\$\\
\poptabs}
\end{lexlist}
\stopx

Several aspects of the output shown above are worth noting.  
Initially, the program assumes that the hearer has no specific
knowledge of any particular objects in the knowledge base.  Note
however, that every proposition put forth by the generator is assumed
to be incorporated into the hearer's set of beliefs.  Consequently,
the descriptive phrase ``an audio journal,'' which is new information
in the first paragraph, is omitted from the second.  Additionally, when
presenting the proposition `{\em Audiofad is an audio journal},' the
generator is able to recognize the similarity with the corresponding
proposition about Stereofool (i.e. both propositions are abstractions
over the single variable open proposition `{\em X is an audio
journal}').  The program therefore interjects the {\em other} property
and produces ``another audio journal.''

Several aspects of the contrastive intonational effects in these
examples also deserve attention.  Because of the content generator's
use of the rhetorical contrast predicate, items are eligible to
receive stress in order to convey contrast before the contrasting
items are even mentioned.  This phenomenon is clearly illustrated by
the clause ``PRAISED by STEREOFOOL'' in \ref{ex:out1}, which is
contrastively stressed before ``REVILED by AUDIOFAD'' is uttered.
Such situations are produced only when the contrasting propositions
are gathered by the content planner in a single invocation of the
generator and identified as contrastive when the rhetorical predicates
are applied.  Moreover, unlike systems that rely solely on word class
and given/new distinctions for determining accentual patterns, the
system is able to produce contrastive accents on pronouns despite
their ``given'' status, as shown in~\ref{ex:out2}.

\section{Conclusions}

The generation architecture described above and implemented in Quintus 
Prolog produces paragraph-length, spoken monologues
concerning objects in a simple knowledge base.  
The architecture relies on a mapping between a two-tiered information
structure representation and intonational tunes to produce speech that
makes appropriate contrastive distinctions prosodically.
The process of natural
language generation, in accordance with much of the recent literature in
the field, is divided into three processes: high-level content planning,
sentence planning, and surface generation.  Two points concerning the role
of intonation in the generation process are emphasized.  First, since
intonational phrasing is dependent on the division of utterances into theme
and rheme, and since this division relates consecutive sentences to one
another, matters of information structure (and hence intonational phrasing)
must be largely resolved during the high-level planning phase.  Second,
since accentual decisions are made with respect to the particular
linguistic realizations of discourse properties and entities (e.g. the
choice of referring expressions), these matters cannot be fully resolved
until the sentence planning phase.

\section{Acknowledgments}

The author is grateful for the advice and helpful suggestions of Mark
Steedman, Justine Cassell, Kathy McKeown, Aravind Joshi, Ellen Prince,
Mark Liberman, Matthew Stone, Beryl Hoffman and Kris
Th\'{o}risson as well as the anonymous ACL reviewers.  Without the AT\&T
Bell Laboratories TTS system, and the patient advice on its use from
Julia Hirschberg and Richard Sproat, this work would not have been possible.
This research was funded by NSF grants IRI91-17110 and IRI95-04372
and the generous sponsors of the MIT Media Laboratory.

\end{document}